# Semantic Similarity Based on Corpus Statistics and Lexical Taxonomy

Jay J. Jiang
Department of Management Sciences
University of Waterloo
Waterloo, Ontario, Canada N2L 3G1
jjiang@uwaterloo.ca

David W. Conrath
MGD School of Business
McMaster University
Hamilton, Ontario, Canada L8S 4M4
conrathd@mcmaster.ca

**Abstract**

This paper presents a new approach for measuring semantic similarity/distance between words and concepts. It combines a lexical taxonomy structure with corpus statistical information so that the semantic distance between nodes in the semantic space constructed by the taxonomy can be better quantified with the computational evidence derived from a distributional analysis of corpus data. Specifically, the proposed measure is a combined approach that inherits the edge-based approach of the edge counting scheme, which is then enhanced by the node-based approach of the information content calculation. When tested on a common data set of word pair similarity ratings, the proposed approach outperforms other computational models. It gives the highest correlation value ($r = 0.828$) with a benchmark based on human similarity judgements, whereas an upper bound ($r = 0.885$) is observed when human subjects replicate the same task.

## 1. Introduction

The characteristics of polysemy and synonymy that exist in words of natural language have always been a challenge in the fields of Natural Language Processing (NLP) and Information Retrieval (IR). In many cases, humans have little difficulty in determining the intended meaning of an ambiguous word, while it is extremely difficult to replicate this process computationally. For many tasks in psycholinguistics and NLP, a job is often decomposed to the requirement of resolving the semantic relation between words or concepts. One needs to come up with a consistent computational model to assess this type of relation. When a word level semantic relation requires exploration, there are many potential types of relations that can be considered: hierarchical (e.g. IS-A or hypernym-hyponym, part-whole, etc.), associative (e.g. cause-effect), equivalence (synonymy), etc. Among these, the hierarchical relation represents the major and most important type, and has been widely studied and applied as it maps well to the human cognitive view of classification (i.e. taxonomy). The IS-A relation, in particular, is a typical representative of the hierarchical relation. It has been suggested and employed to study a special case of semantic relations — semantic similarity or semantic distance (Rada et al. 1989). In this study of semantic similarity, we will take this view, although it excludes some potential useful information that could be derived from other relations.

The study of words/terms relationships can be viewed in terms of the information sources used. The least information used are knowledge-free approaches that rely exclusively on the corpus data themselves. Under the corpus-based approach, word relationships are often derived from their co-occurrence distribution in a corpus (Church and Hanks 1989, Hindle 1990, Grefenstette 1992). With the introduction of machine readable dictionaries, lexicons, thesauri, and taxonomies, these manually built pseudo-knowledge bases provide a natural framework for organising words or concepts into a semantic space. Kozima and Furugori (1993) measured word distance by adaptive scaling of a vector space generated from LDOCE (*Longman Dictionary of Contemporary English*). Morris and Hirst (1991) used Roget's thesaurus to detect word semantic relationships. With the recently developed lexical taxonomy WordNet (Miller 1990, Miller et al. 1990), many researches have taken the advantage of this broad-coverage taxonomy to study word/concept relationships (Resnik 1995, Richardson and Smeaton 1995).

In this paper, we will discuss the use of the corpus-based method in conjunction with lexical taxonomies to calculate semantic similarity between words/concepts. In the next section we will describe the thread and major methods in modelling semantic similarity. Based on the discussion, we will present a new similarity measure, which is a combined approach of previous methods. In section 3, experiments are conducted to evaluate various computational models compared against human similarity judgements. Finally, we discuss the related work and future direction of this study.

## 2. Semantic Similarity in a Taxonomy

There are certain advantages in the work of semantic association discovery by combining a taxonomy structure with corpus statistics. The incorporation of a manually built pseudo-knowledge base (e.g. thesaurus or taxonomy) may complement the statistical approach where "true" understanding of the text is unobtainable. By doing this, the statistics model can take advantage of a conceptual space structured by a hand-crafted taxonomy, while providing computational evidence from manoeuvring in the conceptual space via distributional analysis of corpora data. In other words, calculating the semantic association can be transformed to the estimation of the conceptual similarity (or distance) between nodes (words or concepts) in the conceptual space generated by the taxonomy. Ideally, this kind of knowledge base should be reasonably broad-coverage, well structured, and easily manipulated in order to derive desired associative or similarity information.

Since a taxonomy is often represented as a hierarchical structure, which can be seen as a special case of network structure, evaluating semantic similarity between nodes in the network can make use of the structural information embedded in the network. There are several ways to determine the conceptual similarity of two words in a hierarchical semantic network. Topographically, this can be categorised as node based and edge based approaches, which correspond to the information content approach and the conceptual distance approach, respectively.

### 2.1. Node-based (Information Content) Approach

One node based approach to determine the conceptual similarity is called the information content approach (Resnik 1992, 1995). Given a multidimensional space upon which a node represents a unique concept consisting of a certain amount of information, and an edge represents a direct association between two concepts, the similarity between two concepts is the extent to which they share information in common. Considering this in a hierarchical concept/class space, this common information "carrier" can be identified as a specific concept node that subsumes both of the two in the hierarchy. More precisely, this super-class should be the first class upward in this hierarchy that subsumes both classes. The similarity value is defined as the information content value of this specific super-ordinate class. The value of the information content of a class is then obtained by estimating the probability of occurrence of this class in a large text corpus.

Following the notation in information theory, the information content (IC) of a concept/class $c$ can be quantified as follows:

$$IC(c) = \log^{-1} P(c), \qquad (1)$$

where $P(c)$ is the probability of encountering an instance of concept $c$. In the case of the hierarchical structure, where a concept in the hierarchy subsumes those lower in the hierarchy, this implies that $P(c)$ is monotonic as one moves up the hierarchy. As the node's probability increases, its information content or its informativeness decreases. If there is a unique top node in the hierarchy, then its probability is 1, hence its information content is 0.

Given the monotonic feature of the information content value, the similarity of two concepts can be formally defined as:

$$sim(c_1, c_2) = \max_{c \in Sup(c_1, c_2)} [IC(c)] = \max_{c \in Sup(c_1, c_2)} [-\log p(c)], \qquad (2)$$

where $Sup(c_1, c_2)$ is the set of concepts that subsume both $c_1$ and $c_2$. To maximize the representativeness, the similarity value is the information content value of the node whose IC value is the largest among those super classes. In another word, this node is the "lowest upper bound" among those that subsume both $c_1$ and $c_2$.

In the case of multiple inheritances, where words can have more than one sense and hence multiple direct super classes, word similarity can be determined by the best similarity value among all the class pairs which their various senses belong to:

$$sim(w_1, w_2) = \max_{c_1 \in sen(w_1)\ c_2 \in sen(w_2)} [sim(c_1, c_2)], \qquad (3)$$

where $sen(w)$ denotes the set of possible senses for word $w$.

For the implementation of the information content model, there are some slightly different approaches toward calculating the concept/class probabilities in a corpus. Before giving the detailed calculation, we need to define two concept sets: *words(c)* and *classes(w)*. *Words(c)* is the set of words subsumed (directly or indirectly) by the class $c$. This can be seen as a sub-tree in the whole hierarchy, including the sub-tree root *c*. *Classes(w)* is defined as the classes in which the word *w* is contained; in another word, it is the set of possible senses that the word *w* has:

$$classes(w) = \{c | w \in words(c)\}. \quad (4)$$

Resnik (1995) defined a simple class/concept frequency formula:

$$freq(c) = \sum_{w \in words(c)} freq(w). \quad (5)$$

Richardson and Smeaton (1995) proposed a slightly different calculation by considering the number of word senses factor:

$$freq(c) = \sum_{w \in words(c)} \frac{freq(w)}{|classes(w)|} \quad (6)$$

Finally, the class/concept probability can be computed using maximum likelihood estimation (MLE):

$$P(C) = \frac{freq(c)}{N} \quad (7)$$

This methodology can be best illustrated by examples. Assume that we want to determine the similarities between the following classes: *(car, bicycle)* and *(car, fork)*. Figure 1 depicts the fragment of the WordNet (Version 1.5) noun hierarchy that contains these classes. The number in the bracket of a node indicates the corresponding information content value. From the figure we find that the similarity between *car* and *bicycle* is the information content value of the class *vehicle*, which has the maximum value among all the classes that subsume both of the two classes, i.e. *sim(car, bicycle)* = 8.30. In contrast, *sim(car, fork)* = 3.53. These results conform to our perception that cars and forks are less similar than cars and bicycles.

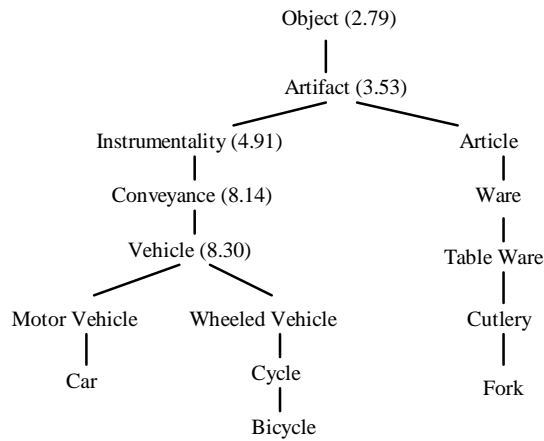

Figure 1. Fragments of the WordNet noun taxonomy

## 2.2. Edge-based (Distance) Approach

The edge based approach is a more natural and direct way of evaluating semantic similarity in a taxonomy. It estimates the distance (e.g. edge length) between nodes which correspond to the concepts/classes being compared. Given the multidimensional concept space, the conceptual distance can conveniently be measured by the geometric distance between the nodes representing the concepts. Obviously, the shorter the path from one node to the other, the more similar they are.

For a hierarchical taxonomy, Rada et al. (1989) pointed out that the distance should satisfy the properties of a metric, namely: zero property, symmetric property, positive property, and triangular inequality. Furthermore, in an IS-A semantic network, the simplest form of determining the distance between two elemental concept nodes, A and B, is the shortest path that links A and B, *i.e.* the minimum number of edges that separate A and B (Rada et al. 1989).

In a more realistic scenario, the distances between any two adjacent nodes are not necessarily equal. It is therefore necessary to consider that the edge connecting the two nodes should be weighted. To determine the edge weight automatically, certain aspects should be considered in the implementation. Most of these are typically related to the structural characteristics of a hierarchical network. Some conceivable features are: local network density (the number of child links that span out from a parent node), depth of a node in the hierarchy, type of link, and finally, perhaps the most important of all, the strength of an edge link. We will briefly discuss the concept for each feature:

- With regard to network density, it can be observed that the densities in different parts of the hierarchy are higher than others. For example, in the *plant/flora* section of WordNet the hierarchy is very dense. One parent node can have up to several hundred child nodes. Since the overall semantic mass is of a certain amount for a given node (and its subordinates), the local density effect (Richardson and Smeaton 1995) would suggest that the greater the density, the closer the distance between the nodes (*i.e.* parent child nodes or sibling nodes).

- As for node depth, it can be argued that the distance shrinks as one descends the hierarchy, since differentiation is based on finer and finer details.

- Type of link can be viewed as the relation type between nodes. In many thesaurus networks the hyponym/hypernym (IS-A) link is the most common concern. Many edge-based models consider only the IS-A link hierarchy (Rada et al. 1989, Lee et al. 1993). In fact, other link types/relations, such as Meronym/Holonym (Part-of, Substance-of), should also be considered as they would have different effects in calculating the edge weight, provided that the data about the type of relation are available.

- To differentiate the weights of edges connecting a node and all its child nodes, one needs to consider the link strength of each specific child link. This could be measured by the closeness between a specific child node and its parent node, against those of its siblings. Obviously, various methods could be applied here. In particular, this is the place where corpus statistics could contribute. Ideally the method chosen should be both theoretical sound and computational efficient.

Two studies have been conducted in edge-based similarity determination by responding to the above concerns. Richardson and Smeaton (1995) considered the first two and the last factors in their edge weight calculation for each link type. Network density is simply counting the number of edges of that type. The link strength is a function of a node's information content value, and those of its siblings and parent nodes. The result of these two operations is then normalised by dividing them by the link depth. Notice that the precise formula of their implementation was not given in the paper.

Sussna (1993) considered the first three factors in the edge weight determination scheme. The weight between two nodes $c_1$ and $c_2$ is calculated as follows:

$$wt(c_1, c_2) = \frac{wt(c_1 \rightarrow_r c_2) + wt(c_2 \rightarrow_{r'} c_1)}{2d} \qquad (8)$$

given

$$wt(x \rightarrow_r y) = \max_r - \frac{\max_r - \min_r}{n_r(x)} \qquad (9)$$

where $\rightarrow_r$ is a relation of type r, $\rightarrow_{r'}$ is its reverse, $d$ is the depth of the deeper one of the two, *max* and *min* are the maximum and minimum weights possible for a specific relation type *r* respectively, and $n_r(x)$ is the number of relations of type *r* leaving node *x*.

Applying this distance formula to a word sense disambiguation task, Sussna (1993) showed an improvement where multiple sense words have been disambiguated by finding the combination of senses from a set of contiguous terms which minimizes total pairwise distance between senses. He found that the performance is robust under a number of perturbations; however, depth factor scaling and restricting the type of link to a strictly hierarchical relation do noticeably impair performance.

In determining the overall edge based similarity, most methods just simply sum up all the edge weights along the shortest path. To convert the distance measure to a similarity measure, one may simply subtract the path length from the maximum possible path length (Resnik 1995):

$$sim(w_1, w_2) = 2d_{max} - [\min_{\substack{c_1 \in sen(w_1) \\ c_2 \in sen(w_2)}} len(c_1, c_2)], \qquad (10)$$

where $d_{max}$ is the maximum depth of the taxonomy, and the *len* function is the simple calculation of the shortest path length (*i.e.* weight = 1 for each edge).

## 2.3. Comparison of the Two Approaches

The two approaches target semantic similarity from quite different angles. The edge-based distance method is more intuitive, while the node-based information content approach is more theoretically sound. Both have inherent strength and weakness.

Rada et al. (1989) applied the distance method to a medical domain, and found that the distance function simulated well human assessments of conceptual distance. However, Richardson and Smeaton (1995) had concerns that the measure was less accurate than expected when applied to a comparatively broad domain (e.g. WordNet taxonomy). They found that irregular densities of links between concepts result in unexpected conceptual distance outcomes. Also, without causing serious side effects elsewhere, the depth scaling factor does not adjust the overall measure well due to the general structure of the taxonomy (e.g. higher sections tend to be too similar to each other).

In addition, we feel that the distance measure is highly depended upon the subjectively pre-defined network hierarchy. Since the original purpose of the design of the WordNet was not for similarity computation purpose, some local network layer constructions may not be suitable for the direct distance manipulation.

The information content method requires less information on the detailed structure of a taxonomy. It is not sensitive to the problem of varying link types (Resnik 1995). However, it is still dependent on the skeleton structure of the taxonomy. Just because it ignores information on the structure it has its weaknesses. It normally generates a coarse result for the comparison of concepts. In particular, it does not differentiate the similarity values of any pair of concepts in a sub-hierarchy as long as their "smallest common denominator" (i.e. the lowest super-ordinate class) is the same. For example, given the concepts in Figure 1, the results of the similarity evaluation between *(bicycle, table ware)* and *(bicycle, fork)* would be the same. Also, other type of link relations information is overlooked here. Additionally, in the calculation of information content, polysemous words will have an exaggerated content value if only word (not its sense) frequency data are used (Richardson and Smeaton 1995).

## 2.4. A Combined Approach

We propose a combined model that is derived from the edge-based notion by adding the information content as a decision factor. We will consider various concerns of the edge weighting schemes discussed in the previous section. In particular, attention is given to the determination of the link strength of an edge that links a parent node to a child node.

We first consider the link strength factor. We argue that the strength of a child link is proportional to the conditional probability of encountering an instance of the child concept $c_i$ given an instance of its parent concept $p$: $P(c_i | p)$.

$$P(c_i|p) = \frac{P(c_i \cap p)}{P(p)} = \frac{P(c_i)}{P(p)} \tag{11}$$

Notice that the definition and determination of the information content (see equations 1 and 5) indicate that $c_i$ is a subset of $p$ when a concept's informativeness is concerned. Following the standard argument of information theory, we define the link strength (LS) by taking the negative logarithm of the above probability. We obtain the following formula:

$$LS(c_i, p) = -\log(P(c_i|p)) = IC(c_i) - IC(p). \tag{12}$$

This states that the link strength (LS) is simply the difference of the information content values between a child concept and its parent concept.

Considering other factors, such as local density, node depth, and link type, the overall edge weight (wt) for a child node $c$ and its parent node $p$ can be determined as follows:

$$wt(c, p) = \left(\beta + (1-\beta)\frac{\overline{E}}{E(p)}\right)\left(\frac{d(p)+1}{d(p)}\right)^\alpha [IC(c) - IC(p)] \, T(c, p), \tag{13}$$

where $d(p)$ denotes the depth of the node $p$ in the hierarchy, $E(p)$ the number of edges in the child links (i.e. local density), $\overline{E}$ the average density in the whole hierarchy, and $T(c,p)$ the link relation/type factor. The parameters $\alpha$ ($\alpha \geq 0$) and $\beta$ ($0 \leq \beta \leq 1$) control the degree of how much the node depth and density factors contribute to the edge weighting computation. For instance, these contributions become less significant when $\alpha$ approaches 0 and $\beta$ approaches 1.

The overall distance between two nodes would thus be the summation of edge weights along the shortest path linking two nodes.

$$Dist(w_1, w_2) = \sum_{c \in \{path(c_1, c_2) - LSuper(c_1, c_2)\}} wt(c, parent(c)) \tag{14}$$

where $c_1 = sen(w_1)$, $c_2 = sen(w_2)$, and *path* ($c_1$, $c_2$) is the set that contains all the nodes in the shortest path from $c_1$ to $c_2$. One of the elements of the set is $LSuper(c_1, c_2)$, which denotes the lowest super-ordinate of $c_1$ and $c_2$. In the special case when only link strength is considered in the weighting scheme of equation 13, *i.e.* $\alpha = 0$, $\beta = 1$, and $T(c,p) = 1$, the distance function can be simplified as follows:

$$Dist(w_1, w_2) = IC(c_1) + IC(c_2) - 2 \times IC(LSuper(c_1, c_2)) \tag{15}$$

Imagine a special multidimensional semantic space where every node (concept) in the space lies on a specific axis and has a mass (based on its information content or informativeness). The semantic distance between any such two nodes is the difference of their semantic mass if they are on the same axis, or the addition of the two distances calculated from each node to a common node where two axes meet if the two original nodes are on different axes. It is easy to prove that the proposed distance measure also satisfies the properties of a metric.

## 3. Evaluation

### 3.1. Task Description

It would be reasonable to evaluate the performance of machine measurements of semantic similarity between concepts by comparing them with human ratings on the same setting. The simplest way to implement this is to set up an experiment to rate the similarity of a set of word pairs, and examine the correlation between human judgement and machine calculations. To make our experimental results comparable with other previous experiments, we decided to use the same sample of 30 noun pairs that were selected in an experiment when only human subjects were involved (Miller and Charles 1991), and in another more recent experiment when some computational models were constructed and compared as well (Resnik 1995). In fact, in the Resnik (1995) experiment, he replicated the human judgements on the same set of word pairs that Miller and Charles did. When the correlation between his replication and the one done by Miller and Charles (1991) was calculated, a baseline from human ratings was obtained for evaluation, which represents an upper bound that one could expect from a machine computation on the same task. In our experiment, we compare the proposed model with the node-based Information Content model developed by Resnik (1995) and the basic edge-based edge counting model, in the context of how well these perform against human ratings (i.e. the upper bound).

For consistency in comparison, we will use semantic similarity measures rather than the semantic distance measures. Hence our proposed distance measure needs to be converted to a similarity measure. Like the edge counting measure in equation 10, the conversion can be made by subtracting the total edge weights from the maximum possible total edge weights. Note that this conversion does not affect the result of the evaluation, since a linear transformation of each datum will not change the magnitude of the resulting correlation coefficient, although its sign may change from positive to negative.

### 3.2. Implementation

The noun portion of the latest version (1.5) of WordNet was selected as the taxonomy to compute the similarity between concepts. It contains about 60,000 nodes (synsets). The frequencies of concepts were estimated using noun frequencies from a universal semantic concordance SemCor (Miller et al. 1993), a semantically tagged text consisting of 100 passages from the Brown Corpus. Since the tagging scheme was based on the WordNet word sense definition, this enables us to obtain a precise frequency distribution for each node (synset) in the taxonomy. Therefore it avoids potentially spurious results in occasions when only word (not word sense) frequencies are used (Resnik 1995). The downside of using the SemCor data is the relatively small size of the corpus due to the need to manually tag the sense for each word in the corpus. Slightly over 25% of the WordNet noun senses actually appeared in the corpus. Nevertheless, this is the only publicly available sense tagged corpus. The MLE method would seem unsuitable for probability estimation from the SemCor corpus. To circumvent the problem of data sparseness, we use the Good-Turing estimation with linear interpolation.

### 3.3. Results

Table 1 lists the complete results of each similarity rating measure for each word pair. The data on human ratings are from the publication of previous results (Miller and Charles 1991, Resnik 1995). Notice that two values in Resnik's replication are not available, as he dropped two noun pairs in his experiment since the word *woodland* was not yet in the WordNet taxonomy at that time. The correlation values between the similarity ratings and the mean ratings reported by Millers and Charles are listed in Table 2. The optimal parameter settings for the proposed similarity approach are: $\alpha=0.5$, $\beta=0.3$. Table 3 lists the results of the correlation values for the proposed approach given a combination of a range of parameter settings.

| Word Pair | | M&C means | Replication means | $Sim_{edge}$ | $Sim_{node}$ | $Sim_{dist}$ |
|---|---|---|---|---|---|---|
| car | automobile | 3.92 | 3.9 | 30 | 10.358 | 30 |
| gem | jewel | 3.84 | 3.5 | 30 | 17.034 | 30 |
| journey | voyage | 3.84 | 3.5 | 29 | 10.374 | 27.497 |
| boy | lad | 3.76 | 3.5 | 29 | 9.494 | 25.839 |
| coast | shore | 3.7 | 3.5 | 29 | 12.223 | 28.702 |
| asylum | madhouse | 3.61 | 3.6 | 29 | 15.492 | 28.138 |
| magician | wizard | 3.5 | 3.5 | 30 | 14.186 | 30 |
| midday | noon | 3.42 | 3.6 | 30 | 13.558 | 30 |
| furnace | stove | 3.11 | 2.6 | 23 | 3.527 | 17.792 |
| food | fruit | 3.08 | 2.1 | 24 | 2.795 | 23.775 |
| bird | cock | 3.05 | 2.2 | 29 | 9.122 | 26.303 |
| bird | crane | 2.97 | 2.1 | 27 | 9.122 | 24.452 |
| tool | implement | 2.95 | 3.4 | 29 | 8.84 | 29.311 |
| brother | monk | 2.82 | 2.4 | 25 | 2.781 | 19.969 |
| crane | implement | 1.68 | 0.3 | 26 | 4.911 | 19.579 |
| lad | brother | 1.66 | 1.2 | 26 | 2.781 | 20.326 |
| journey | car | 1.16 | 0.7 | 0 | 0 | 17.649 |
| monk | oracle | 1.1 | 0.8 | 23 | 2.781 | 18.611 |
| cemetery | woodland | 0.95 | NA | 0 | 0 | 10.672 |
| food | rooster | 0.89 | 1.1 | 18 | 1.03 | 17.657 |
| coast | hill | 0.87 | 0.7 | 26 | 8.917 | 25.461 |
| forest | graveyard | 0.84 | 0.6 | 0 | 0 | 14.52 |
| shore | woodland | 0.63 | NA | 25 | 2.795 | 16.836 |
| monk | slave | 0.55 | 0.7 | 26 | 2.781 | 20.887 |
| coast | forest | 0.42 | 0.6 | 24 | 2.795 | 15.538 |
| lad | wizard | 0.42 | 0.7 | 26 | 2.781 | 20.717 |
| chord | smile | 0.13 | 0.1 | 20 | 4.452 | 17.535 |
| glass | magician | 0.11 | 0.1 | 22 | 1.03 | 17.098 |
| noon | string | 0.08 | 0 | 0 | 0 | 12.987 |
| rooster | voyage | 0.08 | 0 | 0 | 0 | 12.506 |

Table 1. Word Pair Semantic Similarity Measurement

| Similarity Method | Correlation (r) |
|---|---|
| Human Judgement (replication) | 0.8848 |
| Node Based (Information Content) | 0.7941 |
| Edge Based (Edge Counting) | 0.6004 |
| Combined Distance Model | 0.8282 |

Table 2. Summary of Experimental Results (30 noun pairs)

### 3.4. Discussion

The results of the experiment confirm that the information content approach proposed by Resnik (1995) provides a significant improvement over the traditional edge counting method. It also shows that our proposed combined approach outperforms the information content approach. One should recognize that even a small percentage improvement over the existing approaches is of significance since we are nearing the observed upper bound.

The results from Table 3 conform to our projection that the density factor and the depth factor in the hierarchy do affect (although not significantly) the semantic distance metric. A proper selection of these two factors will enhance the distance estimation. Setting the density factor parameter at $\beta=0.3$ seems optimal as most of the resultant values outperform others under a range of depth factor settings. The optimal depth scaling factor $\alpha$ ranges from 0 to 0.5, which indicates it is less influential than the density factor. This would support the Richardson and Smeaton (1995) argument about the difficulty of the adjustment of the depth scaling factor. Another explanation would be that this factor is already absorbed in the proposed link strength consideration. Overall, there is a small performance improvement (2.1%) over the result when only the link strength factor is considered. Since the results are not very sensitive to the variation in parameter settings, we can conclude that they are not the major determinants of the overall edge weight.

| Depth Factor ($\alpha$) | Density Factor ($\beta$) | | | |
|---|---|---|---|---|
| | $\beta=1.0$ | $\beta=0.5$ | $\beta=0.3$ | $\beta=0.2$ |
| $\alpha=2$ | 0.79844 | 0.81104 | 0.81153 | 0.80658 |
| $\alpha=1$ | 0.80503 | 0.82255 | 0.82625 | 0.82266 |
| $\alpha=0.5$ | 0.80874 | 0.82397 | 0.82817 | 0.82509 |
| $\alpha=0$ | 0.81127 | 0.82284 | 0.82737 | 0.82411 |
| $\alpha=-1$ | 0.81435 | 0.81598 | 0.81818 | 0.81349 |
| $\alpha=-2$ | 0.81315 | 0.80228 | 0.80118 | 0.79492 |

Table 3. Correlation coefficient values of various parameter settings for the proposed approach

Further examinations of the individual results in Table 1 may provide a deeper understanding of the model's performance. The ratings in the table are sorted in descending order based on Miller and Charles (1991) findings. This trend can be observed more or less consistently in four other ratings. However, there are some abnormalities that exist in the results. For example, the pair *'furnace-stove'* was given high similarity values in human ratings, whereas a very low rating (second to the lowest) was found in the proposed distance measure. A further look at their classification in the WordNet hierarchy seems to provide an explanation. Figure 2 depicts a portion of WordNet hierarchy that includes all the senses of these two words. We can observe that *furnace* and *stove* are classified under very distinct substructures. Their closest super-ordinate class is *artifact*, which is a very high level abstraction. It would be more reasonable if the substructure containing *furnace* were placed under the class of *device* or *appliance*. If so the distance between *furnace* and *stove* would have been shorter and closer to humans' judgements. This observation re-enforces our earlier thought that the structure of a taxonomy may generate a bias towards a certain distance calculation due to the nature of its classification scheme.

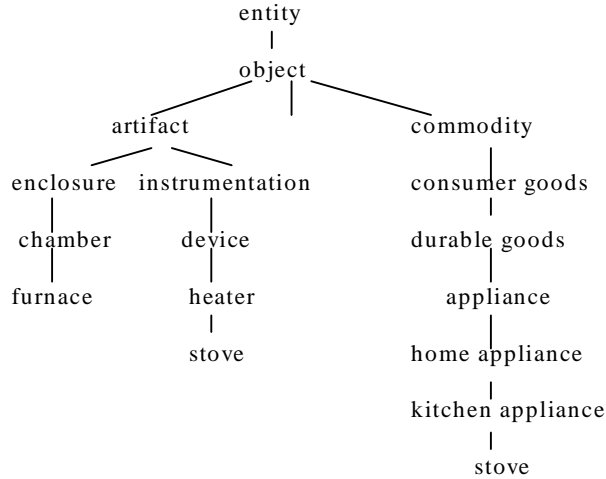

Figure 2. A fragment of WordNet taxonomy

Table 4 shows calculations of the correlation coefficients based on removing the '*furnace-stove*' pair due to a questionable classification of the concept *furnace* in the taxonomy. The result shows an immediate improvement of all the computational models. In particular, our proposed model indicates a large marginal lead.

| Similarity Method | Correlation (r) |
|---|---|
| Node Based (Information Content) | 0.8191 |
| Edge Based (Edge Counting) | 0.6042 |
| Combined Distance Model | 0.8654 |

Table 4. Summary of Experimental Results
(29 noun pairs, removing the '*furnace - stove*' pair)

## 4. Related Work

Closely related works to this study are those that were aligned with the thread of our discussion. In the line of the edge-based approach, Rada et al. (1989) and Lee et al. (1993) derived semantic distance formulas using the edge counting principle, which were then used to support higher level result ranking in document retrieval. Sussna (1993) defined a similarity measure that takes into account taxonomy structure information. Resnik's (1995) information content measure is a typical representative of the node-based approach. Most recently, Richardson and Smeaton (1995) and Smeaton and Quigley (1996) worked on a combined approach that is very similar to ours.

One of the many applications of semantic similarity models is for word sense disambiguation (WSD). Agirre and Rigau (1995) proposed an interesting conceptual density concept for WSD. Given the WordNet as the structured hierarchical network, the conceptual density for a sense of a word is proportional to the number of contextual words that appear on a sub-hierarchy of the WordNet where that particular sense exists. The correct sense can be identified as the one that has the highest density value.

Using an online dictionary, Niwa and Nitta (1994) built a reference network of words where a word as a node in the network is connected to other words that are its definitional words. The network is used to measure the conceptual distance between words. A word vector is defined as the list of distances from a word to a certain set of selected words. These selected words are not necessarily its definitional words, but rather certain types of representational words called *origins*. Word similarity can then be computed by means of their distance vectors. They compared this proposed dictionary-based distance vector method with a corpus-based co-occurrence vector method for WSD and found the latter has a higher precision performance. However, in a test of leaning positive or negative meanings from example words, the former gave remarkable higher precision than the latter. Kozima and Furugori (1993) also proposed a word similarity measure by spreading activation on a semantic net composed by the online dictionary LDOCE.

In the area of IR using NLP, approaches have be pursued to take advantage of the statistical term association results (Strzalkowski and Vauthey 1992, Grefenstette 1992). Typically, the text is first parsed to generated syntactic constructs. Then the *head-modifier* pairs are identified for various syntactical structure. Finally, a specific term association algorithm (similar to the mutual information principle) is applied to the comparison process on a single term/concept basis. Although only modest improvement has been shown, the significance of this approach is that it does not require any domain-specific knowledge or the sophisticated NLP techniques. In essence, our proposed combination model is similar to this approach, except that we also resort to extra knowledge sources—machine readable lexical taxonomies.

## 5. Conclusion

In this paper, we have presented a new approach for measuring semantic similarity between words and concepts. It combines the lexical taxonomy structure with corpus statistical information so that the semantic distance between nodes in the semantic space constructed by the taxonomy can be better quantified with the computational evidence derived from distributional analysis of corpus data. Specifically, the proposed measure is a combined approach that inherits the edge-based approach of the edge counting scheme, which is enhanced by the node-based approach of information content calculation. When tested on a common data set of word pair similarity ratings, the proposed approach outperforms other computational models. It gives the highest correlation value ($r=0.828$), with a benchmark resulting from human similarity judgements, whereas an upper bound ($r=0.885$) is observed when human subjects are replicating the same task.

One obvious application of this approach is for word sense disambiguation. In fact, this is part of the ongoing work. Further applications would be in the field of information retrieval. With the lesson learned from Richardson and Smeaton (1995), when they applied their similarity measure to free text document retrieval, it seems that the IR task would benefit most from the semantic similarity measures when both document and query are relatively short in length (Smeaton and Quigley 1996).